Cross-modality (CT-MRI) prior augmented deep learning for robust lung tumor segmentation from small MR datasets

Jue Jiang[1], Yu-Chi Hu[1], Neelam Tyagi[1], Pengpeng Zhang[1], Andreas Rimner[2], Joseph O. Deasy[1], Harini Veeraraghavan[1*]

*[1]Department of Medical Physics, Memorial Sloan-Kettering Cancer Center, New York, NY 10065, USA*

*[2]Department of Radiation Oncology, Memorial Sloan-Kettering Cancer Center, New York, NY 10065, USA*

*corresponding author
Address: Box 84 - Medical Physics, Memorial Sloan Kettering Cancer Center, 1275 York Avenue, New York, NY 10065
Phone: +1-646-888-8015
Email: veerarah@mskcc.org

**Purpose:** Accurate tumor segmentation is a requirement for magnetic resonance (MR)-based radiotherapy. Lack of large expert annotated MR datasets makes training deep learning models difficult. Therefore, a cross-modality (MR-CT) deep learning segmentation approach that augments training data using pseudo MR images produced by transforming expert-segmented CT images was developed.

**Methods:** Eighty-One T2-weighted MRI scans from 28 patients with non-small cell lung cancers (9 with pre-treatment and weekly MRI and the remainder with pre-treatment MRI scans) were analyzed. Cross-modality prior encoding the transformation of CT to pseudo MR images resembling T2w MRI was learned as a generative adversarial deep learning model. This model augmented training data arising from 6 expert-segmented T2w MR patient scans with 377 pseudo MRI from non-small cell lung cancer CT patient scans with obtained from the Cancer Imaging Archive. A two-dimensional Unet implemented with batch normalization was trained to segment the tumors from T2w MRI. This method was benchmarked against (a) standard data augmentation and two state-of-the art cross-modality pseudo MR-based augmentation and (b) two segmentation networks. Segmentation accuracy was computed using Dice similarity coefficient (DSC), Hausdroff distance metrics, and volume ratio.

**Results:** The proposed approach produced the lowest statistical variability in the intensity distribution between pseudo and T2w MR images measured as Kullback-Leibler divergence of 0.069. This method produced the highest segmentation accuracy with a DSC of (0.75 ± 0.12) and the lowest Hausdroff distance of (9.36 mm ± 6.00mm) on the test dataset. This approach produced highly similar estimations of tumor growth as an expert (P = 0.37).

**Conclusions:** A novel deep learning MR segmentation was developed that overcomes the limitation of learning robust models from small datasets by leveraging learned cross-modality priors to augment training. The results show the feasibility of the approach and the corresponding improvement over the state-of-the-art methods.

## 1. INTRODUCTION

High-dose radiation therapy that is delivered over a few fractions is now a standard of care for lung tumors. The ability to accurately target tumors will enable clinicians to escalate the dose delivered to tumors while minimizing the dose delivered to normal structures. Tumor delineation remains the weakest link in achieving highly accurate precision radiation treatments using computed tomography (CT) as a treatment planning modality [1, 2]. The superior soft tissue contrast of magnetic resonance (MR) images facilitates better visualization of tumor and adjacent normal tissues, especially for those cancers located close to the mediastinum. Such improved visualization makes MR an attractive modality for radiation therapy [3]. Therefore, fast and accurate tumor segmentation on magnetic resonance imaging (MRI) could help to deliver high-dose radiation while reducing treatment complications to normal structures.

Deep convolutional network-based learning methods are the best-known techniques for segmentation and have been expected to achieve human expert-level performance in image analysis applications in radiation oncology [4]. However, such methods require a tremendous amount of data to train models that are composed of a very large number of parameters. Because MRI is not the standard of care for thoracic imaging, it is difficult to obtain enough MR image sets with expert delineations to train a deep learning method. Deep learning with a large number of MRI for segmenting lung tumors is difficult due to (i) lack of sufficient number of training data as MRI is not standard of care for lung imaging, and (ii) lack of sufficient number of expert delineated contours required for training.

The goal of this work is to compute a robust deep learning model for segmenting tumors from MRI despite lacking sufficiently large expert-segmented MRIs (N > 100 cases) for training.

This work addresses the challenge of learning from small expert-segmented MR datasets by developing and testing a novel cross-modality deep domain adaptation approach that employs a model encoding the transformation of CT into an image representation resembling MRI as prior knowledge to augment training a deep learning network with few expert-segmented T2w MR datasets. This prior knowledge, encoded as a deep generative adversarial network (GAN) [5, 6] transforms expert-segmented CT into pseudo MR images with expert segmentations. The pseudo MR images resemble real MR by mimicking the statistical intensity variations. To overcome the issues of the existing methods, which cannot accurately model the anatomical characteristics of atypical structures, including tumors [7], a tumor-attention loss that regularizes the GAN model and produces pseudo MR with well-preserved tumor structures is introduced. Figure 1 shows example pseudo MR images generated from a representation CT image using the state-of-the-art cycle GAN[8] and the proposed approach. The corresponding T2w MRI for the CT image acquired within a week is also shown alongside for comparison.

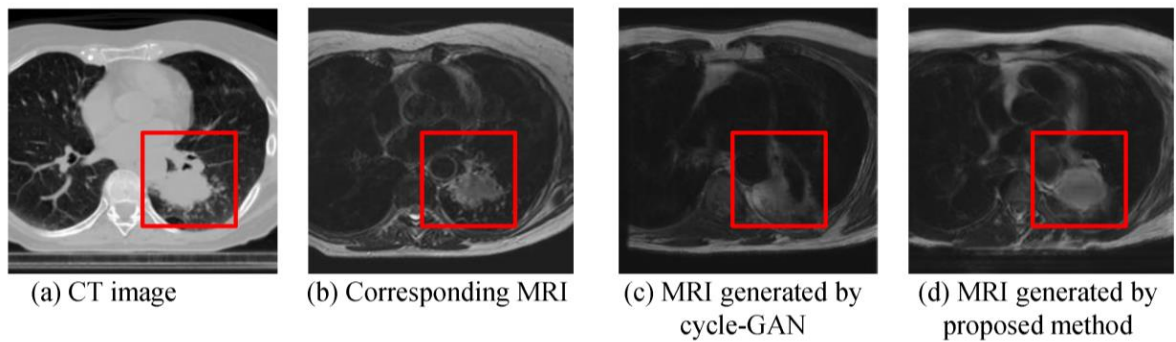

| (a) CT image | (b) Corresponding MRI | (c) MRI generated by cycle-GAN | (d) MRI generated by proposed method |

Figure 1. Pseudo MR image synthesized from a representative (A) CT image using (C) cycle-GAN [8] and (D) proposed method. The corresponding T2w MRI scan for (A) is shown in (B).

This paper is a significant extension of our work in [9] that introduced the tumor-aware loss to use pseudo MRI from CT for MRI segmentation. Instead of the standard Unet [10]. Unet with batch normalization in all layers was implemented to standardize the feature maps produced at all layers. The utility of cross-modality augmentation was evaluated with two different segmentation architectures including the residual fully convolutional networks (residual FCN) [11] and dense fully convolutional networks (dense FCN) [12]. Finally, a subset of patients who had serial weekly imaging during treatment were analyzed to assess the feasibility of this approach for longitudinal tumor segmentation.

## 2. MATERIALS AND METHODS

### 2.1 Patient and image characteristics

A total of 81 T2-weighted (T2W) MRIs from 28 patients enrolled in a prospective IRB-approved study with non-small lung cancer (NSCLC) scanned on a 3T Philips Ingenia scanner (Medical Systems, Best, Netherlands) before and every week during conventional fractionated external beam radiotherapy of 60 Gy were analyzed. Nine out of the 28 patients underwent weekly MRI scans during treatment for up to seven MRI scans. The tumor sizes ranged from 0.28cc to 264.37 cc with average of 50.87 ±55.89 cc. Respiratory triggered two-dimensional (WD) axial T2W turbo spin-echo sequence MRIs were acquired using a 16-element phased array anterior coil and a 44-element posterior coil and the following scan parameters: TR/TE = 3000-6000/120 ms, slice thickness = 2.5 mm and in-plane resolution of 1.1 x 0.97 mm2 flip angle = 90°, number of slices = 43, number of signal averages = 2, and field of view = 300 x 222 x 150 mm3. Radiation oncologist delineated tumor contours served as ground truth. In addition, CT images with expert delineated contours from an unrelated cohort of 377 patients with NSCLC [13] available from The Cancer Imaging Archive (TCIA) [14] were used for training.

## 2.2 Approach overview

A two-step approach, as shown in Figure 1, consisting of (a) Step 1: CT to pseudo MR expert-segmented dataset transformation, and (b) Step 2: MR segmentation combining expert-segmented T2W MRI with pseudo MRI produced from Step 1 was employed. Pseudo MR image synthesis from CT was accomplished through cross-modality domain adaption (Figure 1a) using expert-segmented CTs with unrelated T2w MR images without any expert annotation. The approach consists of two simultaneously trained GANs that produce pseudo MR and pseudo CT from CT and MR images, respectively. The MR segmentation network in Step 2 consists of a standard 2D Unet (Figure 1b) with layer-wise batch normalization applied to standardize the image features from the training set. Figure 1(c-e) depicts the losses used for pseudo MR synthesis whereas Figure 1f corresponds to the segmentation loss.

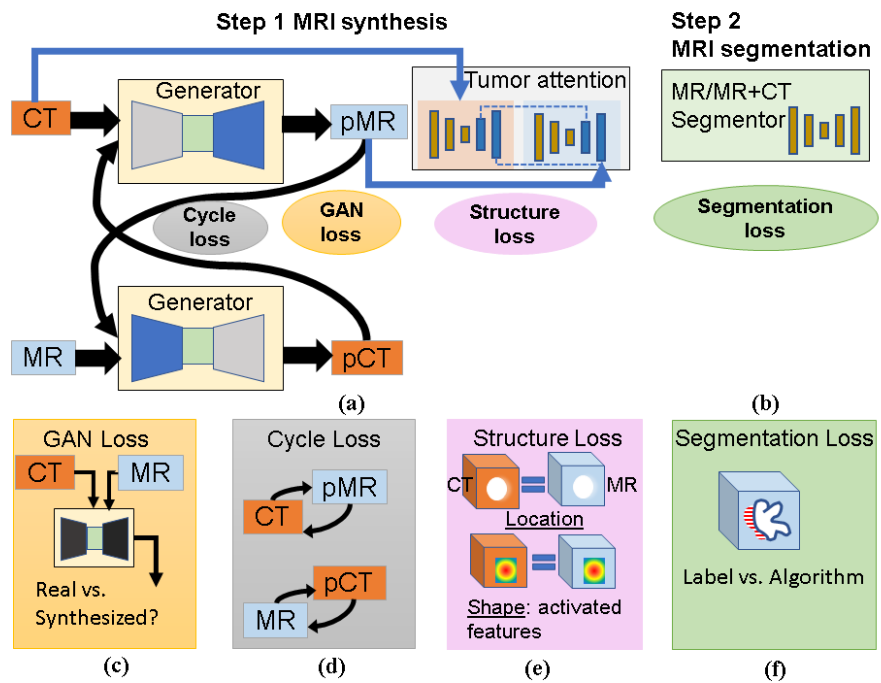

Figure 2 Approach overview. (a) Pseudo MR synthesis, (b) MR segmentation training using pseudo MR with T2w MR. Visual description of losses used in training the networks in (a) and (b), namely, (c) GAN or adversarial loss, (d) cycle or cycle consistency loss, (e) tumor-attention loss enforced using structure and shape loss, and (f) segmentation loss computed using Dice overlap coefficient.

## 2.3 Cross-modality augmented deep learning segmentation:

**2.3.1 Image pre-processing:** Prior to analysis using the deep networks, all T2W MR images were standardized to remove patient-dependent signal intensity variations using the method in [15]. The CT HU and MRI intensity values were clipped to (-1000, 500) and (0,667), respectively to force the networks to model the intensity range encompassing thoracic anatomy. The resulting clipped images were normalized to the range (-1,1) to ensure numerical stability in the deep networks that employ *tanh* functions.

Table 1 provides a brief description of the notation used in the paper for the developed method.

Table 1: List of notations

| Notation | Description |
|---|---|
| $\{X_{CT}, y_{CT}\}$ | CT images and expert segmentations |
| $\{X_{MR}, y_{MR}\}$ | MR images and expert segmentations |
| pMR | pseudo MR image |
| pCT | pseudo CT image |
| $G_{CT \rightarrow MR}$ | Generator for producing pMR images |
| $G_{MR \rightarrow CT}$ | Generator for producing pCT images |
| $D_{CT \rightarrow MR}$ | Discriminator to distinguish pMR from real T2w MR images |
| $D_{MR \rightarrow CT}$ | Discriminator to distinguish pCT from real CT images |
| $\{L_{adv}, L_{adv}^{MR}, L_{adv}^{CT}\}$ | Total adversarial loss, adversarial loss for pMRI and for pCT |
| $\{L_{cyc}, L_{cyc}^{MR}, L_{cyc}^{CT}\}$ | Total cycle loss, cycle loss for pMR and pCT images |
| $L_{tumor}^{shape}$ | Tumor shape loss |
| $L_{tumor}^{loc}$ | Tumor location loss |

### 2.3.2 Step 1: Cross-modality prior (MR-CT) learning for pseudo MR generation:

The first step in the approach is to learn a cross-modality prior to produce pseudo MR from CT scans. The cross-modality prior is learned as a generational adversarial network (GAN) model [5], which extracts a model of the anatomical characteristics of tissues in MR and CT using unlabeled MRI and expert-segmented CT scans.

Two GAN networks are trained simultaneously to generate pseudo MRI and pseudo CT from CT and MRI, respectively. Each GAN is composed of a generator and a discriminator that work together in a minmax fashion, whereby the generator learns to produce images that confound the discriminator's classification. The discriminator trains to correctly distinguish between real and synthesized images. This loss is called the adversarial loss $L_{adv}$ (Figure 1[c]) and is computed as a sum of losses for generating pseudo MRI and pseudo CT as: $L_{adv} = L_{adv}^{MR} + L_{adv}^{CT}$, shown in eq (1) and (2).

$$L_{adv}^{MR} = E_{x_m \sim X_{MRI}}[\log(D_{MRI}(x_m))] + E_{x_c \sim X_{CT}}[\log(1 - D_{MRI}(G_{CT \rightarrow MRI}(x_c)))] \qquad (1)$$

$$L_{adv}^{CT} = E_{x_c \sim X_{CT}}[\log(D_{CT}(x_c))] + E_{x_{mri} \sim X_{MRI}}[\log(1 - D_{CT}(G_{MRI \rightarrow CT}(x_m)))] \qquad (2)$$

Cycle loss, $L_{cyc}$ introduced in [8] was used to soften the requirement of perfectly aligned CT and MRI from the same patients for training using adversarial loss, thereby, enabling the methods to use imaging modalities from unrelated patients to produce the transformations. $L_{cyc}$ forces the networks to preserve spatial correspondences between the pseudo and original modalities (see Figure 1 [d]) and is computed as, $L_{cyc} = L_{cyc}^{MR} + L_{cyc}^{CT}$ for synthesizing pseudo MR and pseudo CT images, as shown in eq (3).

$$L_{cyc} = E_{x_c \sim X_{CT}}\left[\left|\left|(G_{CT \rightarrow MRI}(x_m^{'}) - x_c)\right|\right|\right] + E_{x_m \sim X_{MRI}}[|||(G_{MRI \rightarrow CT}(x_c^{'}) - x_m||] \qquad (3)$$

Where $x_c^{'}$ and $x_m^{'}$ are the generated pseudo CT and pseudo MRI from generator $G_{MRI \rightarrow CT}$ and $G_{CT \rightarrow MRI}$, respectively. Note that pseudo CTs are simply a by-product of generating the pseudo MR and are generated to add additional regularization in synthesizing pseudo MR images.

Adversarial and cycle consistency losses enable a network to only learn global and frequently occurring structures as a result of which atypical or structures with lots of inter-patient variations such as tumors are lost upon transformation from CT to pseudo MR images. This problem is addressed by introducing a structure-specific, tumor attention loss, $L_{tumor}$.

The tumor attention loss is implemented using a pair of 2D Unets, which is composed of 5 convolutional and 5 deconvolutional layers with skip connections and concatenation between them. The two Unets are trained to produce a coarse tumor segmentation using the pseudo MR and CT images. The two networks share the last two layers to produce identical segmentation, thereby, constraining the generated pseudo MR to preserve tumors in the same location as the CT image. It is composed of tumor shape loss, $L_{tumor}^{shape}$ (eq (4)) and tumor location loss, $L_{tumor}^{loc}$ (eq (5)).

The tumor shape loss minimizes the differences in the feature map activations (through Euclidean distances) from the penultimate network layer whereas the tumor location loss forces the two networks to produce identical tumor segmentations (through DSC).

$$L_{tumor}^{shape} = \frac{1}{C \times W \times H} ||\emptyset_{CT}(x_c) - \emptyset_{CT}(G_{CT \to MRI}(x_c))||^2 \qquad (4)$$

$$L_{tumor}^{loc} = E_{x_c \sim X_{CT}}[y_c|G_{CT \to MRI}(x_c)] + E_{x_c \sim X_{CT}}[y_c|x_c] \qquad (5)$$

where C,W and H are the feature channel, width and height of the feature.

The total loss is computed as:

$$L_{total} = L_{adv} + \lambda_{cyc}L_{cyc} + \lambda_{shape}L_{tumor}^{shape} + \lambda_{loc}L_{tumor}^{loc} \qquad (6)$$

where $\lambda_{cyc}$, $\lambda_{shape}$ and $\lambda_{loc}$ are the weighting coefficients for each loss. The computed losses are back-propagated to update the weights of the networks to produce a pseudo MRI and pseudo CT from the two networks. The algorithm for pseudo MR generation is shown in Algorithm 1.

---

Algorithm 1: CT to pseudo MRI generation

Input : CT modality: Xc and label yc, T2w MR modality: $X_m$

Output: pMRI image $X_m^{'}$ and label $X_m^{'} = y$

1 $\theta_G, \theta_D, \theta_T \leftarrow initialize$

2 for Iters ≤ Max Iter do

3 $\quad X_c, X_m \leftarrow images\ sampled\ minibatch\ from\ CT\ and\ MRI$

4 $\quad Calculate\ L_{adv}, L_{cyc}, L_{tumor}^{shape}, L_{tumor}^{loc}$ by equation (1)-(2), (3), (4), (5)

5 $\quad \theta_G \overset{+}{\leftarrow} -\Delta_{\theta_G}(L_{adv} + \lambda_{cyc}L_{cyc} + \lambda_{shape}L_{tumor}^{shape} + \lambda_{loc}L_{tumor}^{loc})$

6 $\quad$ Calculate $Ladv$ by equation (1) and (2)

7 $\quad \theta_D \overset{+}{\leftarrow} -\Delta_{\theta_D}L_{adv}$

8 $\quad Calculate\ L_{tumor}^{shape}, L_{tumor}^{loc}$ by equation (4) and (5)

9 $\quad \theta_T \overset{+}{\leftarrow} -\Delta_{\theta_T}(\lambda_{shape}L_{tumor}^{shape} + \lambda_{loc}L_{tumor}^{loc})$

10 end

---

### 2.3.3 Step 2: MRI tumor segmentation

The pseudo MRI dataset, produced from Step 1, is combined with the available expert-segmented T2W MRI scans (N = 6) to train a MRI tumor segmentation network. The standard implementation consists of a 2D Unet [10] architecture which was modified with batch normalization implemented after each convolution operation to standardize the features computed at the different layers. Feature standardization was implemented to help with improved robustness to variations in MR signal intensities among patients. This step results in tumor segmentation from MRI.

## 2.4 Benchmarking segmentation performance

The method's performance was benchmarked by analyzing the impact of training with the learned pseudo-MRI using Unet and two state-of-the-art cross-modality synthesis approaches using the cycle-GAN[8] and masked cycle-GAN [16] methods. The masked cycle-GAN method uses expert-segmentations on both modalities as an additional input channel to specifically focus training on the regions containing tumors.

Besides, two state-of-the-art segmentation architectures called the fully convolutional residual network, or Residual fully convolutional network (FCN), [11] and densely connected FCN, or Dense-FCN[12], were also implemented to verify the utility of proposed data augmentation approach for boosting MRI segmentation.

The residual FCN combines feature maps computed from the second, third, and fourth feature pooling layers through element-wise summation to maintain fixed-size feature maps. Residual FCN was implemented by combining the features extracted from the layers following second, third and fourth pooling operations in the ResNet50 through element-wise summation. Finer image details required for extracting the segmentation is learned by merging features computed from multiple image resolutions. The features from multiple image resolutions are combined by upsampling features back to the image resolution that produces feature maps at the original image resolution (see Figure 3(B)).

The dense FCN successively concatenates feature maps computed from previous layers, thereby increasing the size of the feature maps. A dense feature block is produced by iterative summation of previous feature maps within that block. As features computed from all image resolutions starting from the original image resolution to the lowest resolution are iteratively concatenated, features at all levels are utilized for segmentation. Such a connection also enables the network to implement an implicit dense supervision to better train the features required for the analysis (see Figure 3 (C)).

Differences in the network architectures for the Unet, Residual FCN and Dense-FCN are shown in Figure 3.

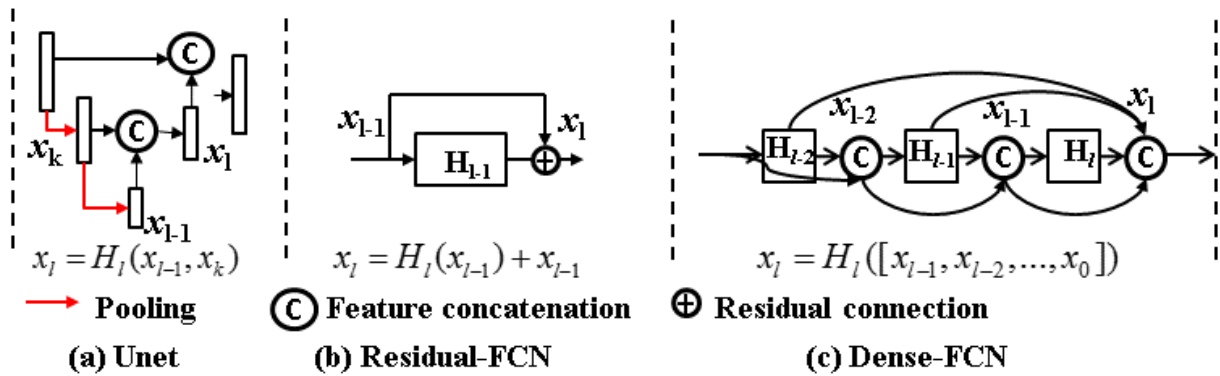

Figure 3  The segmentation architectures, consisting of (A) Unet, (B) Dense-FCN and (c) Residual FCN.

## 2.5 Implementations and training

The GAN networks are implemented in a similar fashion to [9] with two stride-2 convolutions, nine residual blocks, and two fractionally stride convolutions with one-half stride. The PatchGAN [17] discriminator was used by the discriminator to distinguish real from pseudo images. The GANs resulted in labeled pseudo MR images produced by transforming expert-segmented CT scans.

All networks were trained using 2D image patches of size $256 \times 256$ pixels computed from image regions enclosing the tumors. Cross-modality synthesis networks were trained using 32000 CT images, and 9696 unlabeled T2W MR image patches were obtained by standard data augmentation techniques[18], including rotation, stretching, and using different regions of interests containing tumors and normal structures. Segmentation training was performed using 1536 MR images obtained from pre-treatment MRI from 6 consecutive patients and the 32000 synthesized MR images. The best segmentation model was selected from a validation set consisting of image slices from 36 weekly MRI scans not used for model training. Additional testing was performed on a total of 39 MRI scans obtained from 22 patients who were not used in training or validation. Three of those patients contained longitudinal MRI scans (7,7,6).

 All the networks were implemented in the PyTorch library [19] and trained on Nvidia GTX 1080Ti of 12 GB memory with a batch size of 1, during image transfer, and a batch size of 10, during semi-supervised segmentation using Unet, residual-FCN, and dense-FCN. The ADAM algorithm[20], with an initial learning rate of 1e-4, was used during training with preset parameters $\lambda_{cyc}$=10, $\lambda_{shape}$=5 and $\lambda_{loc}$=1.

All the code for CT to MRI transformation as well as the evaluated segmentation networks will be made available for use by other researchers through Github.

## 2.6 Evaluation Metrics

The similarity of pseudo MRI to the T2w MRI was evaluated using Kullback–Leibler divergence[21] (K-L divergence), that quantifies the average statistical differences in the intensity distributions. The K-L divergence was computed using the intensity values within the tumor regions by comparing the intensity histogram for all the generated pseudo MR images and the intensity histogram computed from the training T2w MR images. The KL divergence measure quantifies the similarity in the overall intensity variations between the pseudo MR and the real MR images within the structure of interest, namely, tumors. The K-L divergence is calculated by eq. 2.

$$D_{KL}(P_{sMRI} \parallel Q_{rMRI}) = \sum P_{sMRI} \ln \frac{P_{sMRI}}{Q_{rMRI}} \qquad (7)$$

where $P_{sMRI}$ and $Q_{rMRI}$ indicate the tumor distribution in pseudo MR and T2w MR images and the summation is computed over a fixed number of discretized intensity levels (N = 1000).

The segmentation accuracy was computed using DSC and Hausdorff distance. The DSC is calculated between the algorithm and expert segmentation as:

$$DSC = \frac{2TP}{FP + 2TP + FN} \qquad (8)$$

where, TP is the number of true positives, FP is the number of false positives, and FN is the number of false negatives.

The Hausdorff distance is defined as:

$$Haus(P,T) = \max\{\sup_{p \in S_P} \inf_{t \in S_T} d(p,t), \sup_{t \in S_T} \inf_{p \in S_P} d(t,p)\} \qquad (9)$$

where, P and T are expert and segmented volumes, and p and t are points on P and T, respectively. $S_p$ and $S_t$ correspond to the surface of P and T, respectively. To remove the influence of noise during evaluation, Hausdorff Distance (95%) was used, as recommended by Menze[22]. We also calculated the relative volume ratio, computed as $VR = \frac{|V_{as} - V_{gs}|}{V_{gs}}$, where $V_{as}$ is the volume calculated by algorithm while $V_{gs}$ is the volume calculated by manually segmentation.

Finally, for a small set of patients who had longitudinal follow ups, the tumor growth trends were computed using the Theil-Sen estimator[23], which measures the median of slope (or tumor growth) computed from consecutive time points as:

$$l(\delta_v) = median \lim_{l_i < l_k} (\frac{l(v_k) - l(v_i)}{t_k - t_i}) \tag{10}$$

where $v_k$, $v_i$ are the tumor volumes at times $k$ and $i$ for a lesion $l$.

The difference in growth rate between algorithm and the expert delineation was computed using the Student's T-test.

## 3. RESULTS

### 3.1 Impact of the proposed tumor-attention loss in CT to MRI transformation

Figure 4 shows MRI produced using the proposed (Figure 4d), cycle-GAN (Figure 4b) and masked cycle-GAN (Figure 4c) for two representative examples. The expert-segmented tumor contour placed on the CT (Figure 4a) is shown at the corresponding locations on the pseudo MRI. As shown, this method produced the best preservation of tumors on the pseudo MR images.

As shown in Figure 4 e, this method produced the closest approximation of the distribution of MR single intensities as the T2w MRI. The algorithm generated tumor distribution was computed within the tumors of the 377 pseudo MRI produced from CT, while the T2w MR distribution was obtained from tumor regions from expert-segmented T2w MR used in validation (N=36). This was confirmed on quantitative evaluation wherein, our method resulted in the lowest KL divergence of 0.069 between the pseudo MRI and T2w MRI compared with both the cycle-GAN (1.69) and the masked cycle-GAN (0.32) with 1000 bins to obtain a sufficiently large discretized distribution of intensities.

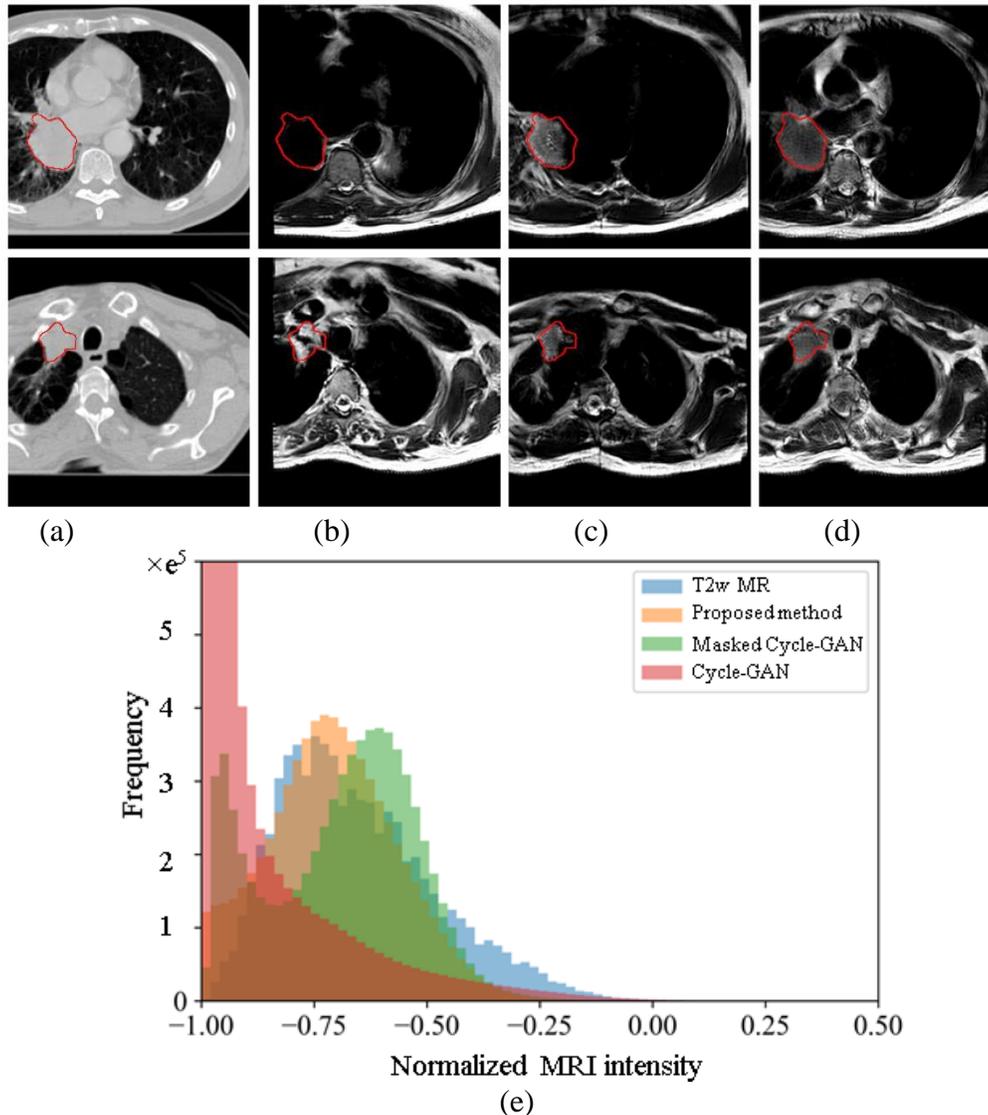

(a)        (b)        (c)        (d)

(e)

Figure 4: CT to pseudo MRI transformation using the analyzed methods. (a) Original CT; pseudo MR image produced using (b) CycleGAN [8]; (c) masked CycleGAN [16]; and (d) proposed method. (e) shows the normalized intensity variation within the tumor region using various methods. The T2w MR intensity distribution within the tumor regions from the validation patients is also shown for comparison.

## 3.2 Impact of data augmentation using transformed MRI on training Unet-based MRI tumor segmentation

Figure 5 shows segmentation results from five randomly chosen patients computed using the proposed Unet implementation trained with only T2w MRI (Figure 5a), T2w MRI combining cross-modality augmentation using cycle-GAN (Figure 5b), and masked cycle-GAN methods (Figure 5c), and the proposed approach (Figure 5d). Figure 5e shows the results of training with only pseudo MRI produced using proposed method and excluding real T2w MRI. As shown, the algorithm generated contours (yellow) closely approximate expert-segmentation (red) when trained through MR augmentation using the proposed method. The overall segmentation accuracy using DSC and Hausdroff distances for these methods are shown in Table. 1. As shown, this method resulted in the highest DSC and lowest Hausdorff distances on test sets that were not used either in training or model selection.

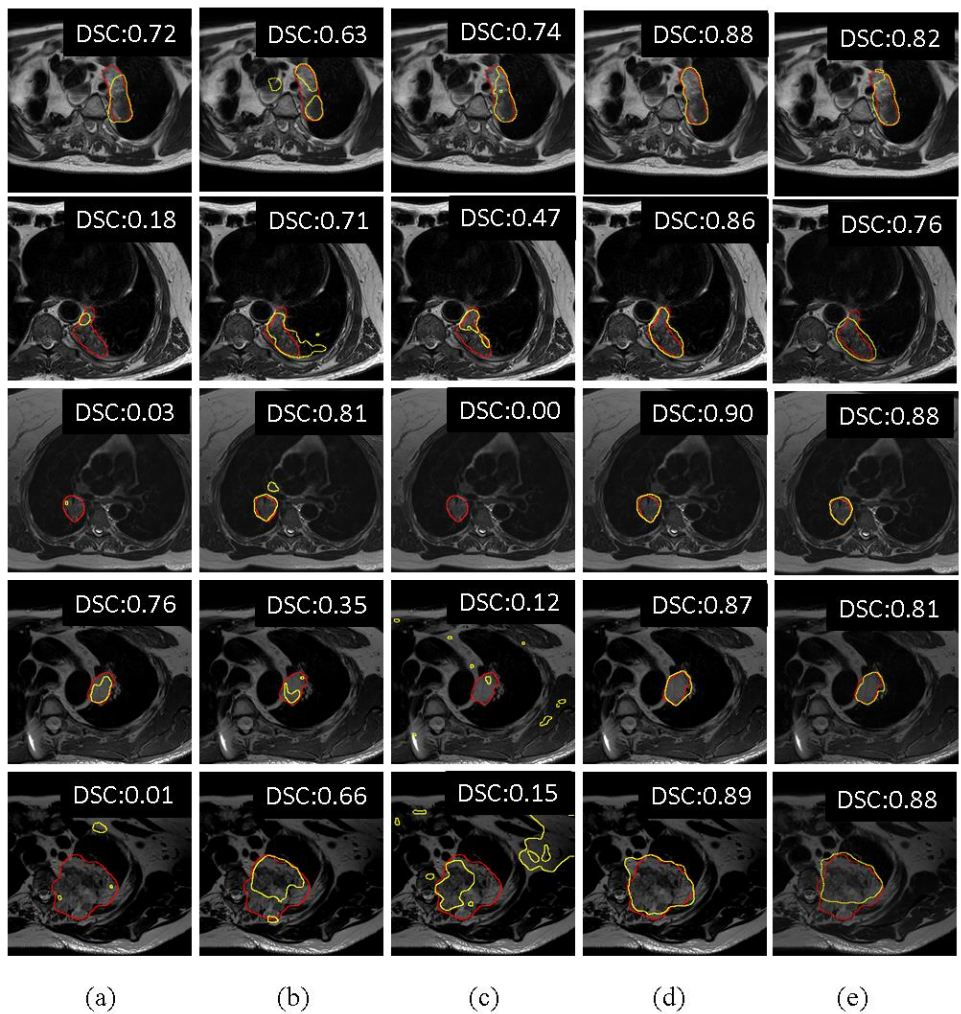

(a)  (b)  (c)  (d)  (e)

Figure 5: Segmentations from representative examples from five different patients using different data augmentation methods. (a) expert-segmented T2w MRI; expert-segmented T2w MRI with pseudo MRI produced using (b) cycle-GAN; (c) masked cycleGAN; (d) proposed method, and (e) pseudo MRI produced using proposed method and excluding expert-segmented T2w MRI. The red contour corresponds to the expert delineation while the yellow contour corresponds to algorithm generated segmentation.

**3.3 Impact of network architecture on tumor segmentation accuracy**

Table II summarizes the segmentation accuracy achieved using three segmentation architectures where training was augmented using five different methods consisting of: expert-segmented T2w MRI, cross-modality augmentation using cycle-GAN, masked cycle-GAN, and the proposed method combined with expert-segmented T2w MRI, and only the pseudo MR images produced using proposed tumor-aware augmentation.

Table II. Segmentation accuracies computed using the various architectures, namely, Unet, residual-FCN, and dense-FCN for the different data augmentation strategies. Segmentation accuracies are shown using Dice similarity coefficient (DSC), 95[th] percentile Hausdroff distance (HD95) and volume ratio (VR) metrics.

Table II segmentation accuracy comparison

| Unet | | | | | | |
|---|---|---|---|---|---|---|
| | Validation | | | Test | | |
| Method | DSC | HD95 mm | VR | DSC | HD95 mm | VR |
| Standard augmentation of T2w MRI | 0.63±0.27 | 7.22±7.19 | 0.34±0.31 | 0.50±0.26 | 18.42±13.02 | 0.54±0.27 |
| cycle-GAN [8] pseudo MRI and T2w MRI | 0.57±0.24 | 11.41±5.57 | 0.50±0.53 | 0.62±0.18 | 15.63±10.23 | 0.54±1.27 |
| masked cycle-GAN [16] pseudo and T2w MR | 0.67±0.21 | 7.78±4.40 | 0.84±0.30 | 0.56±0.26 | 20.77±18.18 | 0.37±0.57 |
| Tumor-aware pseudo MR and T2w MR | **0.70±0.19** | **5.88±2.88** | **0.23±0.15** | **0.75±0.12** | **9.36±6.00** | **0.19±0.15** |
| Tumor-aware pseudo MR | 0.62+0.26 | 7.47+4.66 | 0.35±0.29 | 0.72±0.15 | 12.45±10.87 | 0.25±0.19 |

| Residual-FCN | | | | | | |
|---|---|---|---|---|---|---|
| | Validation | | | Test | | |
| Method | DSC | HD95 mm | VR | DSC | HD95 mm | VR |
| Standard augmentation of T2w MRI | 0.62±0.24 | 9.31±5.30 | 0.33±0.29 | 0.50±0.19 | 23.96±17.00 | 0.48±0.23 |
| cycle-GAN [8] pseudo MRI and T2w MRI | 0.47±0.27 | 11.71±5.99 | 0.41±0.30 | 0.52±0.20 | 17.82±12.11 | 0.54±0.89 |
| masked cycle-GAN [16] pseudo and T2w MR | 0.42±0.28 | 16.01±7.65 | 0.44±0.31 | 0.54±0.25 | 20.36±12.02 | 0.54±1.46 |
| Tumor-aware pseudo MR and T2w MR | **0.70±0.21** | **7.56±4.85** | **0.22±0.16** | **0.72±0.15** | **16.64±5.85** | **0.25±0.22** |
| Tumor-aware pseudo MR | 0.59/0.23 | 10.09/5.88 | 0.36±0.32 | 0.69±0.17 | 18.03/12.91 | 0.27±0.25 |

| Dense-FCN | | | | | | |
|---|---|---|---|---|---|---|
| | Validation | | | Test | | |
| Method | DSC | HD95 mm | VR | DSC | HD95 mm | VR |
| Standard augmentation of T2w MRI | 0.68±0.28 | 8.06±4.89 | 0.29±0.25 | 0.57±0.23 | 24.98±16.51 | 0.51±0.99 |
| cycle-GAN [8] pseudo MRI and T2w MRI | 0.60±0.27 | 10.29±6.22 | 0.34±0.26 | 0.60±0.19 | 23.77±17.61 | 0.35±0.55 |
| masked cycle-GAN [16] pseudo and T2w MR | 0.60±0.23 | 9.85±6.85 | 0.48±0.35 | 0.55±0.24 | 25.11±20.89 | 0.47±0.33 |
| Tumor-aware pseudo MR and T2w MR | **0.68±0.20** | **7.37±5.49** | **0.27±0.28** | **0.73±0.14** | **13.04±6.06** | **0.21±0.19** |
| Tumor-aware pseudo MR | 0.62/0.21 | 11.09/5.02 | 0.42±0.29 | 0.67±0.16 | 20.70±10.57 | 0.22±0.19 |

### 3.3 Comparison of longitudinal segmentations produced using algorithm and expert

Figure 6 shows the estimate of tumor growth computed using the proposed method from three patients with weekly MR scans who were not used for training. This method produced a highly similar estimate of tumor growth as an expert as indicated by the differences in tumor growth computed between algorithm and expert segmentation (week one only: $0.067 \pm 0.012$; cycle-GAN [8]: $0.038 \pm 0.028$; masked cycle-GAN[16]: $0.042 \pm 0.041$; proposed: $0.020 \pm 0.011$).

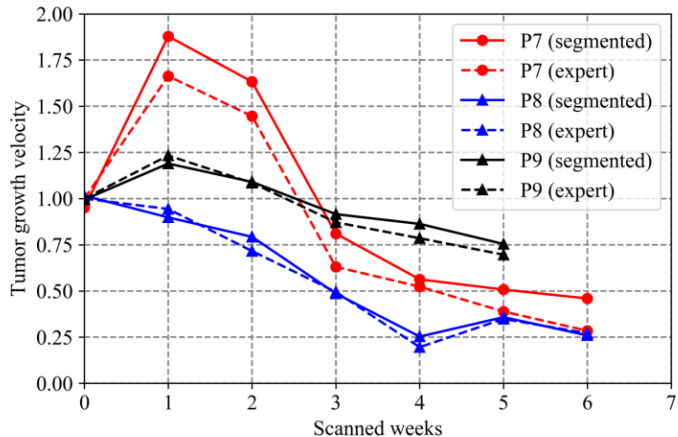

(a)

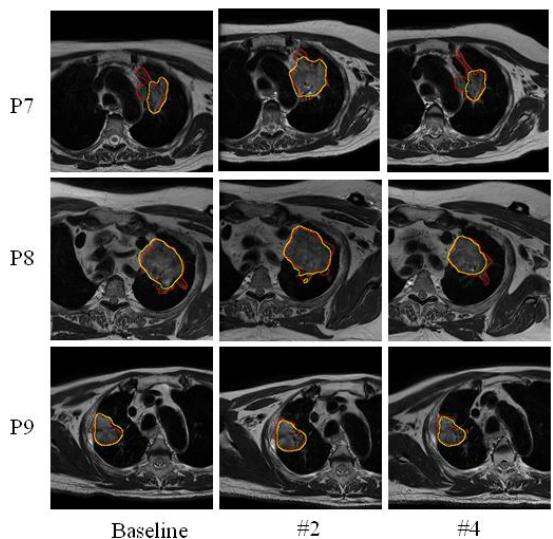

(b)

Figure 6: Longitudinal tumor volumes computed from three example patients using proposed method. (a) Volume growth velocity calculated by proposed versus expert delineation (b) Segmentation results from patient 7 and patient 8. The red contour corresponds to the expert delineation while the yellow contour corresponds to algorithm generated segmentation.

## IV. DISCUSSION

A novel approach for training deep convolutional neural networks (CNNs) for tumor segmentation from MRI with limited expert-segmented training sets was developed by augmenting training data through pseudo MR images obtained from cross-modality transformation of expert-segmented CT datasets from unrelated patients. Our results show that data augmentation obtained through such transformations helps to improve segmentation accuracy on test sets and leads to performance improvements regardless of the chosen segmentation architecture. We found that the Unet using batch normalization applied to each layer was sufficient to achieve highly accurate tumor segmentations as the more advanced residual FCN and dense FCN methods. The batch normalization applied to each layer of Unet served to standardize the feature maps produced from the training images and diminish the patient to patient intensity variations in MR images.

Prior works have used cycle loss [24, 25] to learn the cross-modality prior to transform modalities, where the network learns the transformation by minimizing global statistical intensity differences between the original and a transformed modality (e.g., CT vs. pseudo CT) produced through circular synthesis (e.g., CT to pseudo MR, and pseudo MR to pseudo CT). Although cycle loss eliminates the need for using perfectly aligned CT and MRI acquired from the same patient for training, it cannot model anatomical characteristics of atypical structures, including tumors. In fact, cycle-GAN based methods have been shown to be limited in their ability to transform images even from one MR to a different MR sequence [7]. This approach overcomes the limitations of the prior approaches by incorporating structure-specific losses that constrain the cross-modality transformations to preserve atypical structures including tumors. We note however that all of these methods produce pseudo MR representations that do not create synthetic MR images modeling the same tissue specific intensities as the original MRI. This is intentional as our goal is simply to create an image representation that resembles an MRI and models the global and local (structure of interest) statistical variations that will enable an algorithm to detect and segment tumors. The fact that the pseudo MR images are unlike T2w MRI is demonstrated by the lower accuracy particularly on test sets when excluding any T2w MRI from training. However, it is notable that the pseudo MR images produced by our method still lead to more accurate segmentations compared with the cycle GAN or the highly limited number of T2w MR images.

We speculate that the higher DSC accuracy in the test sets compared to the validation resulted from the higher tumor volumes in validation $31.49\pm42.67$ cc compared with the test datasets $70.86\pm59.80$. It is known that larger tumors can lead to higher DSC accuracies [26].

Unlike most prior works that employ transfer learning to fine tune features learned on other tasks, such as natural images [25,27-28], we chose a data augmentation approach using cross-modality priors. The reason for this is transfer learning performance on networks trained on completely unrelated datasets is easily surpassed even when training from scratch using sparse datasets [29]. This is because features of deeper layers may be too specific for the previously trained task and not easily transferable to a new task. The approach overcomes the issue of learning from small datasets by augmenting training with relevant datasets from a different modality.

This work has a few limitations. First, the approach is limited by the number of test datasets, particularly for longitudinal analysis due to the lack of additional recruitment of patients. Second, due to lack of corresponding patients with CT and MRI, it is not possible to evaluate the patient level correspondences of the pseudo and T2w MR images. Nevertheless, the focus of this work was not to synthesize MR images that are identical to true MRI as is the focus of works using generational adversarial networks to synthesize CT from MR [30]. Instead our goal was to augment the ability of deep learning to learn despite limited training sets by leveraging datasets mimicking intensity variations as the real MR images. To the best of our knowledge, this is one of the first approaches to successfully use cross-domain adaptation learning to augment training data and obtain robust deep CNN models for generating fully automatic and longitudinal segmentation of lung tumors from MRI with access to only a few expert-segmented MRI datasets.

## V. CONCLUSIONS

A novel deep learning approach using cross-modality prior was developed to train robust models for MR lung tumor segmentation from small expert-labeled MR datasets. This method overcomes the limitation of learning robust models from small datasets by leveraging information from expert-segmented CT datasets through a cross-modality prior model. This method surpasses state-of-the-art methods in segmentation accuracy and demonstrates initial feasibility in auto-segmentation of lung tumors from MRI.

## ACKNOWLEDGEMENTS

This work was supported in part by Varian Medical Systems, and partially by the MSK Cancer Center support grant/core grant P30 CA008748, and by NIH R01 CA198121-03. We also thank Dr. Mageras for his insightful suggestions for improving the clarity of the manuscript.

## REFERENCRES

1. Nieh CF. Tumor delineation: the weakest link in the search for accuracy in radiotherapy, *Medical Physics* 2008; 33(4): 136.
2. Eisenhauer E., Therasse P., Bogaerts J. *et al.*, New response evaluation criteria in solid tumours: revised RECIST guideline (version 1.1), *European journal of cancer, 2009,* 45: 228-247.
3. Pollard J M, Wen Z, Sadagopan R, Wang J, Ibbott G S. The future of image-guided radiotherapy will be MR guided. The British journal of radiology 2017; 90(1073): 20160667.
4. Thompson R F, Valdes G, Fuller C D, Carpenter C M, Morin O, Aneja S, *et al.* The Future of Artificial Intelligence in Radiation Oncology. *International Journal of Radiation Oncology• Biology• Physics* 2018; 102(2): 247-248.
5. Goodfellow I, Pouget-Abadie J, Mirza M, *et al.* Generative adversarial nets, in: Advances in Neural Information Processing Systems (NIPS) 2014; p. 2672-2680.
6. Radford A, Metz L, Chintala S. Unsupervised representation learning with deep convolutional generative adversarial networks. *In: International conference in Learning Representations* 2016.
7. J. P. Cohen, M. Luck, and S. Honari, "Distribution Matching Losses Can Hallucinate Features in Medical Image Translation," in: *International Conference on Medical Image Computing and Computer-Assisted Intervention* (MICCAI), Springer, 2018. p. 529-537.

8. Zhu JY, Park T, Isola P, Efros A. Unpaired image-to-image translation using cycle-consistent adversarial networks, in: *Intl. Conf. Computer Vision (ICCV)* 2017: p. 2223-2232.

9. Jiang J., Hu Y.-C., Tyagi N. *et al.*, Tumor-Aware, Adversarial Domain Adaptation from CT to MRI for Lung Cancer Segmentation, in: *International Conference on Medical Image Computing and Computer-Assisted Intervention* (MICCAI, 2018, pp. 777-785.

10. Ronneberger O, Fischer P, Brox T. U-net: Convolutional networks for biomedical image segmentation, in: *International Conference on Medical Image Computing and Computer-Assisted Intervention* (MICCAI), Springer, 2015. p. 234-241.

11. He K, Zhang X, Ren S, Sun J, Deep residual learning for image recognition, *In Proceedings of the IEEE conference on computer vision and pattern recognition* (CVPR), 2016. p. 770-778.

12. Jégou S, Drozdzal M, Vazquez D, Romero A, Bengio Y, The one hundred layers tiramisu: Fully convolutional densenets for semantic segmentation, *in: Computer Vision and Pattern Recognition Workshops* (CVPRW), 2017. p 1175-1183.

13. Aerts HJ, Rios Velazquez E, Leijenaar RT, Parmar C, *et al*. Data from NSCLC-radiomics. The Cancer Imaging Archive, 2015.

14. Clark K, Vendt B, Smith K, Freymann J, *et al*. The cancer imaging archive (TCIA): maintaining and operating a public information repository, *Journal of digital imaging* 2013. 26 (6):1045-1057.

15. Nyúl LG, Udupa JK. On standardizing the MR image intensity scale, Magnetic Resonance in Medicine. 1999 Dec; 42(6): p. 1072-1081.

16. Chartsias A, Joyce T, Dharmakumar R, Tsaftaris SA, Adversarial image synthesis for unpaired multi-modal cardiac data, in: *Intl Workshop on Simulation and Synthesis in Medical Imaging* (ISBI), Springer, 2017. p. 3-13.

17. Isola P, Zhu JY, Zhou T, Efros AA. Image-to-image translation with conditional adversarial networks. in: *Intl. Conf. Computer Vision (ICCV)* 2017: p. 1125-1134

18. Roth HR, Lee CT, Shin HC, Seff A, *et al*. Anatomy-specific classification of medical images using deep convolutional nets. in: Intl Symposium on Biomedical Imaging (ISBI), Springer, 2015.

19. Paszke A, Gross S, Chintala S, Chanan G, *et al*. Automatic differentiation in pytorch.

20. Kingma DP, Ba J. Adam: A method for stochastic optimization. *In: International conference in Learning Representations* 2014 Dec 22.

21. R. O. Duda and P. E. Hart, "Pattern recognition and scene analysis," ed: Wiley, New York, 1973.

22. Menze BH, Jakab A, Bauer S, Kalpathy-Cramer J, Farahani K, Kirby J, Burren Y, Porz N, Slotboom J, Wiest R, Lanczi L, The multimodal brain tumor image segmentation benchmark (BRATS), IEEE Transactions on Medical Imaging 2015. 34(10):1993.

23. Lavagnini I, Badocco D, Pastore P, Magno F. Theil_sen nonparametric regression technique on univariate calibration, inverse regression and detection limits, *Talanta* 2011. 87:180-188.

24. Wolterink JM, Dinkla AM, Savenije MH, Seevinck PR, *et al*. Deep MR to CT synthesis using unpaired data, in: *Intl Workshop on Simulation and Synthesis in Medical Imaging*, Springer, 2017: p. 14-23.

25. Zhang Z, Yang L, Zheng Y. Translating and segmenting multimodal medical volumes with cycle- and shape consistency generative adversarial network, in: *IEEE Conf. on Computer Vision and Pattern Recognition* (CVPR), 2018. P. 9242-9251.

26. G. Sharp, K. D. Fritscher, V. Pekar, M. Peroni, N. Shusharina, H. Veeraraghavan*, et al.*, "Vision 20/20: Perspectives on automated image segmentation for radiotherapy," *Medical physics,* vol. 41, 2014.

27. Long M, Cao Y, Wang J, Jordan MI, Learning transferable features with deep adaptation networks, in: Intl Conf. Machine Learning (ICML), 2015. p. 97-105.

28. Van OA, Ikram MA, Vernooij MW, De Bruijne M. Transfer learning improves supervised image segmentation across imaging protocols, IEEE Transactions on Medical Imaging 2015. 34(5):1018-1030.

29. Wong KCL, Syeda-Mahmood T, Moradi M. Building medical image classifiers with very limited data using segmentation networks. Med Image Analysis, 2018. 49: p. 105-116.

30. H. Emami, M. Dong, S. P. Nejad‐Davarani, and C. Glide‐Hurst, Generating Synthetic CT s from Magnetic Resonance Images using Generative Adversarial Networks, *Medical physics,* 2018,p.3627-3636